\documentclass[final]{cvpr}

\usepackage{times}
\usepackage{epsfig}
\usepackage{graphicx}
\usepackage{amsmath}
\usepackage{amssymb}

\usepackage[pagebackref=true,breaklinks=true,colorlinks,bookmarks=false]{hyperref}
\DeclareMathOperator*{\argmin}{arg\,min}
\usepackage{stmaryrd}
\usepackage{enumitem}

\def\cvprPaperID{0007}
\def\confYear{CVPR 2021}

\begin{document}

\title{Interpreting Expert Annotation Differences in Animal Behavior}

\author{Megan Tjandrasuwita\\
Caltech
\and
Jennifer J. Sun \\
Caltech
\and
Ann Kennedy \\
Northwestern University \\
\AND 
Swarat Chaudhuri \\
UT Austin
\and
Yisong Yue \\
Caltech
\and
}

\maketitle

\begin{abstract}
\vspace{-0.1in}

Hand-annotated data can vary due to factors such as subjective differences, intra-rater variability, and differing annotator expertise. 
We study annotations from different experts who labelled the same behavior classes on a set of animal behavior videos, and observe a variation in annotation styles.
We propose a new method using program synthesis to help interpret annotation differences for behavior analysis.
Our model selects relevant trajectory features and learns a temporal filter as part of a program, which corresponds to estimated importance an annotator places on that feature at each timestamp.
Our experiments on a dataset from behavioral neuroscience demonstrate that compared to baseline approaches, our method is more accurate at capturing annotator labels and learns interpretable temporal filters.
We believe that our method can lead to greater reproducibility of behavior annotations used in scientific studies. We plan to release our code. 
\vspace{-0.1in}

\end{abstract}

\section{Introduction}
Supervised algorithms for animal behavior quantification have become a powerful tool for characterizing the structure of behavior and its regulation by genes and the brain~\cite{kabra2013jaaba,segalin2020mouse,nilsson2020simple,johnson2020probabilistic}. However, different individuals perceive and describe the world in different ways, and this can create significant inter-annotator and inter-lab differences in the behavioral annotations used to construct such supervised classifiers. In image recognition, variability across individuals have been shown to produce different object categorizations~\cite{gomes2012crowdclustering} or labels for the same image data~\cite{lampert2016empirical}.
Similarly, annotator variability has been observed in animal behavior studies, even among experts studying the same behaviors~\cite{leng2020quantitative,segalin2020mouse}. To improve reproducibility and annotator consensus in behavioral experiments, we propose a novel method for automatically generating interpretations of human behavior annotations.

Existing behavior classification models are typically black-box models trained to reproduce human annotations. While these models can achieve high accuracy in the hands of individual labs, it is difficult to interpret differences between models or training sets produced by different individuals~\cite{segalin2020mouse, nilsson2020simple}.
Previous studies have proposed methods for post-hoc interpretation of trained models~\cite{NIPS2017_7062,ribeiro2016should}, but the large number of dimensions and parameters in modern machine learning models can make it difficult to understand how annotators use specific features to annotate behavior.

\begin{figure}
    \centering
    \includegraphics[width=\linewidth]{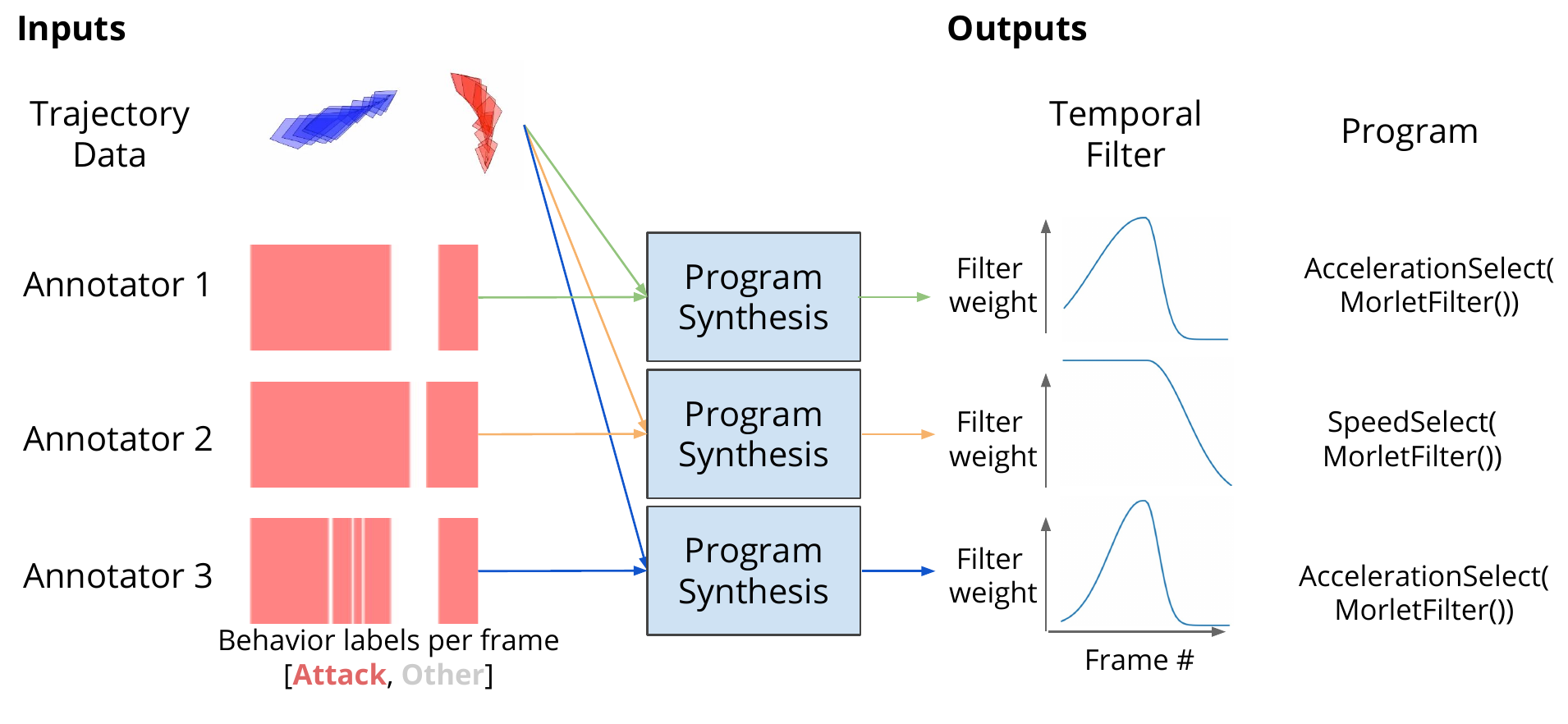}
    \caption{\textbf{Overview.} Given trajectory data and behavior labels, we use program synthesis to learn a programmatic description with temporal filters. These programs can be used to compare differences across annotators.}
    \label{fig:bento_vis}
\end{figure}

To overcome these limitations, we use program synthesis to generate programmatic descriptions from behavior annotations, which can be interpreted without the need for post-hoc analysis.
Program synthesis learns symbolic models from domain-specific languages ~\cite{valkov2018houdini,young2019learning,shah2020learning,ellis2017learning}.
We introduce a domain-specific language for behavior classification, which includes learnable temporal filters and feature selections to identify behaviorally relevant features of animal movement.
We incorporate our setup into an existing program synthesis method~\cite{shah2020learning} to jointly search through the combinatorially large space of program architectures and optimize parameters. 
Our approach produces a program with temporal filters for modeling expert annotations, which domain experts qualitatively found to be interpretable for behavior analysis. Our contributions are: \vspace{-.05in}\begin{itemize}
    \item We combine program synthesis with temporal filtering to generate explanations of behavior annotations.
    \vspace{-0.1in}
    \item We demonstrate our approach on an animal behavior dataset annotated by nine expert annotators.
    \vspace{-0.1in}
    \item We integrate our interpretable programs with an existing tool from domain experts (Bento~\cite{segalin2020mouse}).
\end{itemize}

\section{Related Work}

\textbf{Behavior Modeling.} Automated behavior quantification has enabled scalable analysis of behavioral data in neuroscience and ethology~\cite{kabra2013jaaba,anderson2014toward}. These methods often begin with tracking of anatomically defined keypoints from recorded videos of behaving animals~\cite{segalin2020mouse,mathis2018deeplabcut}. Domain-specific features are then computed from trajectory data and used to train behavior classifiers in the form of neural networks or large random forests~\cite{segalin2020mouse,nilsson2020simple}, which are not easily interpretable. 
Instead, in our approach, we will search through these domain-specific features using program synthesis to produce programs.

\textbf{Interpretable Models.}  Existing interpretability techniques in machine learning generally follow one of two approaches: creating post-hoc explanations of black-box models~\cite{NIPS2017_7062,ribeiro2016should,kim2018interpretability} or learning inherently interpretable models~\cite{lou2012intelligible,shah2020learning,jung2017simple}. We focus on the second approach, and apply techniques in program synthesis~\cite{shah2020learning} to learn a program to model behavior annotations. We compare our method against models with different levels of interpretability, from shallow decision trees to 1-D Convolutional Networks.

We note that there exists a discussion on when machine learning models are interpretable~\cite{rudin2019stop,kaur2020interpreting,lipton2018mythos,lage2019human}. In our work, we focus on our target users, who are domain experts in neuroscience. We work with domain experts to design a DSL which is qualitatively interpretable for them.

\section{Approach}

We consider program learning in the context of sequence classification. We train a program that predicts behavior annotations at each frame from trajectory data, and use this program as a description of an annotator's annotation style.

\subsection{Problem Formulation}
We adopt a problem formulation similar to NEAR~\cite{shah2020learning}. A program is written in a domain-specific language (DSL) and is defined as ($\alpha, \theta$), where $\alpha$ is a discrete program architecture and $\theta$ is a vector of real-valued parameters. We denote the semantics of an architecture by $\llbracket \alpha \rrbracket (x, \theta)$, which is a function parameterized by $\theta$ and applied to input $x$. 

Our goal is to find a program that is both accurate (low prediction error) and interpretable (low structural cost), which we formulate as solving the following optimization problem: 

\begin{equation}
(\alpha^*, \theta^*) = \argmin_{(\alpha, \theta)} (s(\alpha) + \zeta(\alpha, \theta)).
\end{equation}
Here, $\zeta(\alpha, \theta) = \mathbb E_{(x, y) \sim D} [\mathbf{1} (\llbracket \alpha \rrbracket (x, \theta) \neq y)]$ is the standard notion of prediction error.
Since interpretability is a motivating factor, we incentivize short programs by penalizing structural complexity $s(\alpha)$, defined as follows. We let each rule $r$ in our DSL carry a non-negative real cost $s(r)$. The structural cost of an architecture $\alpha$ is $s(\alpha) = \sum_{r \in \mathcal R(\alpha)} s(r)$, where $\mathcal R$ is a multiset of rules used in $\alpha$. 

\textbf{Program Synthesis.} We search over program architectures in a top-down manner. The search is analogous to building a graph $\mathcal G$, where the nodes consists of both partial and complete architectures that are type-consistent with the DSL. The complete architectures are required to be goal nodes. The edges each represent a single-step application of the DSL rules, and are formed between either two partial architectures or a partial and a complete architecture. 

In our approach, we use the program synthesis algorithm NEAR~\cite{shah2020learning}, which learns differentiable programs using an admissible neural heuristic. We note that any program synthesis approach could work within our framework.

\subsection{Learnable Temporal Filters}
    
We develop a DSL from which program synthesis methods can find interpretable programs, based on the Morlet wavelet~\cite{gabor1946theory,grossmann1990reading}. To learn temporal information, our DSL includes a Morlet Filter operation that maps a sequence of vectors to a single vector by taking a weighted sum of the input sequence. The Morlet Filter, denoted by $\psi$, first does a one-to-one mapping between frames $1, \ldots, n$ in the sliding window to values $x_1, \ldots, x_n$, where $x_i \in [-\pi, \pi] \; \forall i=1, \ldots, n$. $\psi$ is then evaluated at each $x_i$ and is defined as:
\begin{align}
    \psi(x; s, w) &= e^{-0.5\left(\frac{x}{(s/w)}\right)^2} \cos(wx), \\
    &\text{where } x \in [-\pi, \pi]. \nonumber
\end{align}

The Morlet Filter is parameterized by $s, w$, where $w$ determines the width of the filter and $s$ controls the wavelet frequency. In our experiments, we use a generalization of the symmetric Morlet Filter by allowing the form of the Morlet Filter to differ between the frames preceding and following the predicted frame. Specifically, the left (preceding) Morlet Filter is parameterized by $s_1, w_1$ whereas the right (following) is parameterized by $s_2, w_2$, resulting in the asymmetric Morlet Filter that we include in our DSL.

Our DSL also includes affine transformations of the following form, where $W$ is a matrix of weights, $x$ is a feature vector, and $b$ is a learned bias:

\begin{equation}
    T(x) = W^T x[i_1, \ldots, i_n] + b.
\end{equation}

Given a full feature vector $x$, the transformation selects a subset of features at indices $i_1, \ldots, i_n$ and applies a simple linear layer to the feature subset. For the purpose of interpretability, we limit $n$ to be 1 or 2, i.e. the transformations either select a single feature, or the two same features for the resident and intruder mice. 

Within our DSL, the Morlet Filter operation is differentiable with respect to parameters $s, w$, allowing the shape of the filter to be discovered through gradient optimization. Similarly, the weights and bias $W, b$ of each affine transformation $T$ are amenable to gradient descent. 

\textbf{Disjunctions.} Our DSL allows \emph{disjunctions} of two or more Morlet Filter operations. The output of a complete Morlet Filter program (the Morlet Filter applied to a sequence of feature vectors, followed by an affine transformation) is a logit. A disjunction combines the predictions of each filter by summing up the outputted logits. In order to reduce variability in the programs found by the disjunction, we perform separate runs of NEAR to find each filter in the disjunction. Once a filter in the disjunction is found, its weights are frozen when discovering the structure and optimizing the parameters of the subsequent filters. This encourages each subsequent filter to explain variance in the dataset that has not been captured by the previous filters.

\section{Experiments}

\subsection{Dataset} 
We use a subset of the MARS~\cite{segalin2020mouse} dataset for studying annotator variability, which consists of ten 10-minute videos at 30Hz of socially interacting mice from a standard resident-intruder assay. These videos are independently annotated for three behaviors of interest by each of nine domain experts. As input, we use a subset of domain-specific features from the MARS dataset: 1) for both mice, we compute head-body angle, body axis ratio, speed, acceleration, tangential velocity, social angle, 2) across mice, we compute area ellipse ratio, whether resident is facing intruder, and minimum distance of resident nose to intruder body.
Here, we consider two binary classification tasks: interact vs. no-interact, and aggression vs. no-aggression. Interaction is defined as frames on which one mouse is sniffing, attacking, or mounting the other; aggression is defined as periods of high-intensity biting, chasing, or grappling.

\subsection{Evaluation Procedure}
We compare the performance of our discovered programs with the following baselines: 1) Decision Trees, a popular choice for both performance and interpretability; 2) 1D Convolutional Neural Networks, a black-box model that is well-suited for processing temporal signals.

\textbf{Decision Trees (DT).} Decision trees are constructed by finding yes/no questions that split the data into the most homogeneous groups. We implement DTs using XGBoost \cite{chen2015xgboost}, a popular framework for training tree ensembles, and test tree classifiers of varying complexity. As input to the decision trees, we pass handcrafted temporal features, produced by convolving our 15 behavior features or their first or second derivatives with a Gaussian filter with standard deviations of 8, 30, or 120 frames This produced 135 total features: 15 original features * 3 derivative orders (0, 1, and 2) * 3 filter widths.

\textbf{1D Convolutional Networks (1D Conv).} In a similar manner to a Morlet Filter, a 1D convolutional neural network produces a weighted sum of a given sequence of vectors- however unlike the Morlet Filter, weights are not constrained to have any specific temporal structure. 
The 1D Conv Net learns a set of weights to convolve with each input feature over time, and the logits from all features are summed for the output predictions.

\begin{figure}
    \centering
    \includegraphics[width=0.8\linewidth]{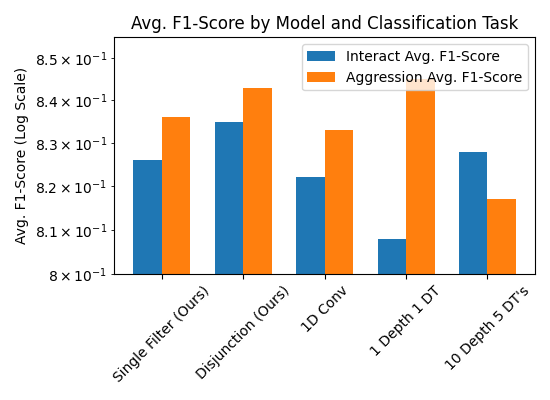}
    \caption{\textbf{Performance of Models Trained on 100\% Training Data.} Bars reflect mean F1 score of each model when trained and tested separately on each of the nine annotators in the MARS dataset.}
    \label{fig:100_bar}
\end{figure}

\textbf{Evaluation Details.} We defined a window of +/- 5 seconds centered about the frame for which behavior was to be predicted, and extracted features of animal poses within this window. We then downsampled data from 30Hz to 6Hz, producing vectors of length 61 for each of the 15 features.

We evaluated all models using the F1-score, defined as the harmonic mean of Precision and Recall. We selected 6 videos for training (106k frames), 2 for validation (40k frames), and 2 for test (39k frames). 
To compare data efficiency, we sub-sampled the training data by randomly sampling trajectories of 1000 frames to achieve desired fractions of the training set size. The sampling also retained a similar class distribution as the full training set. 
For every data fraction (1\%, 10\%, 50\%), we create three different random samples and train all models three times for each sample. The results are reported on the average across these nine repeats, and across the nine annotators.

\subsection{Results}

\def\figsize{0.33}
\def\fighspace{-2mm}
\def\fighspacer{-2mm}
\begin{figure*}
\centering
\begin{tabular}{ccc}
\centering
\scriptsize{Program Filters}\hspace{\fighspacer} & \scriptsize{Decision Trees}\hspace{\fighspacer} &
\scriptsize{1D Conv Net}\hspace{\fighspacer} \\
 \includegraphics[width=\figsize\textwidth]{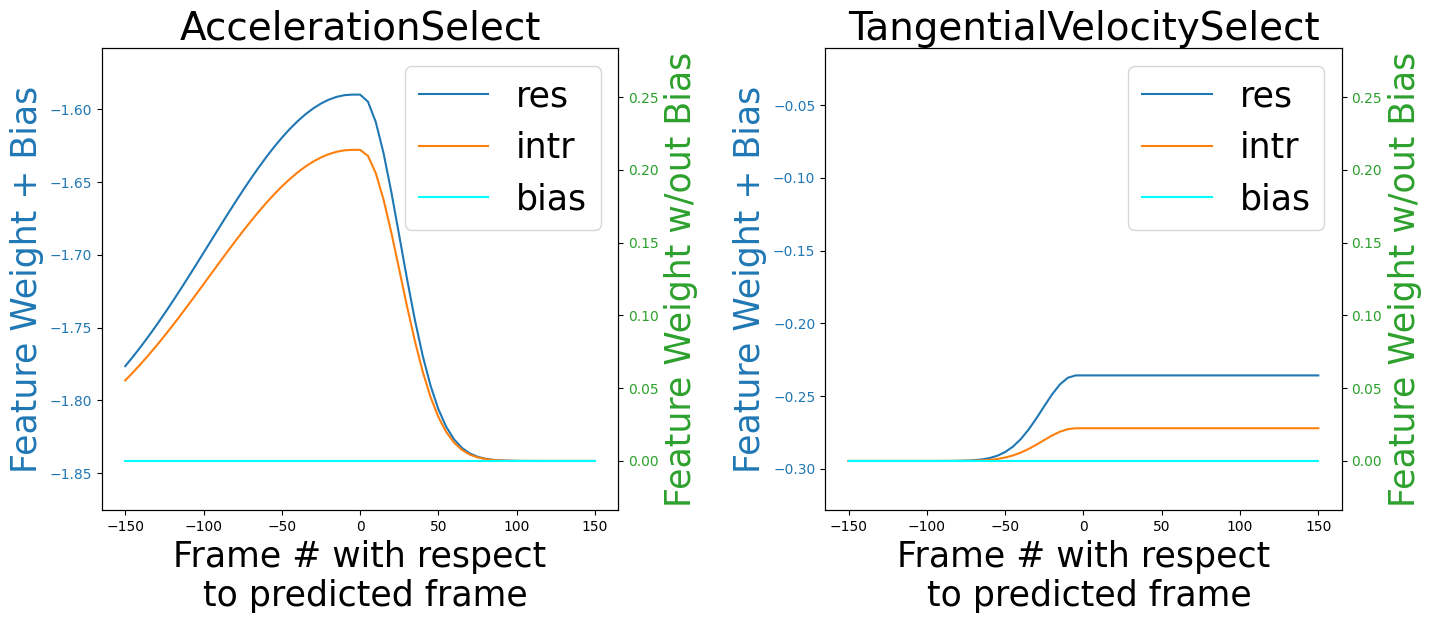}\hspace{\fighspacer} &
 \includegraphics[width=\figsize\textwidth]{{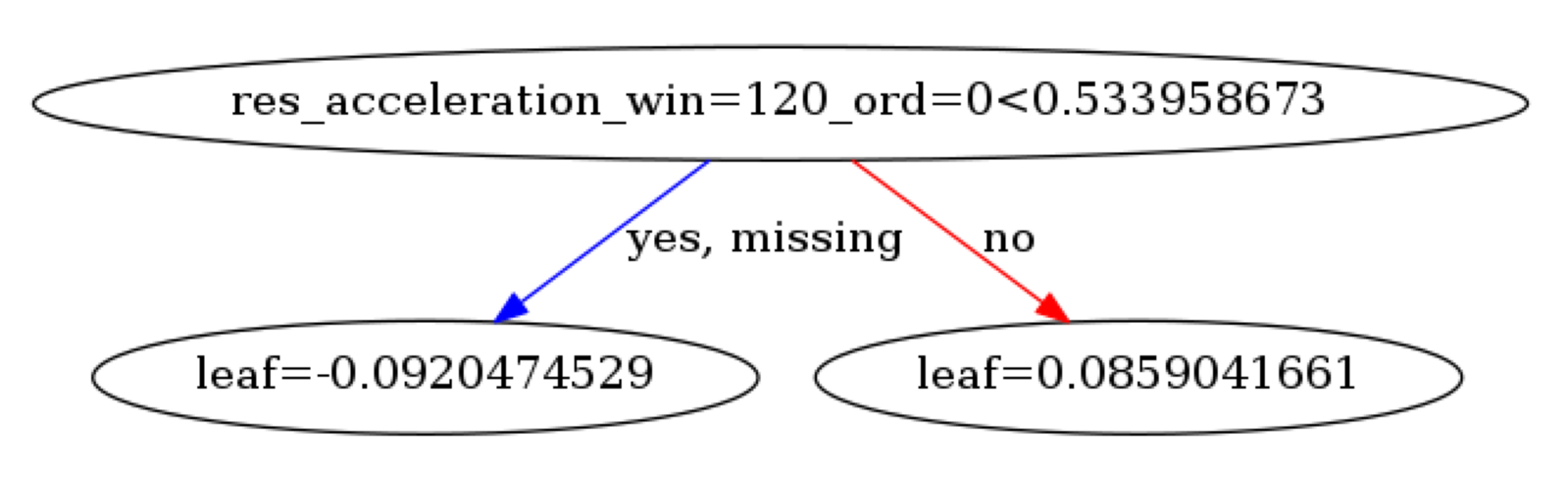}}\hspace{\fighspacer}  &
 \includegraphics[width=\figsize\textwidth]{{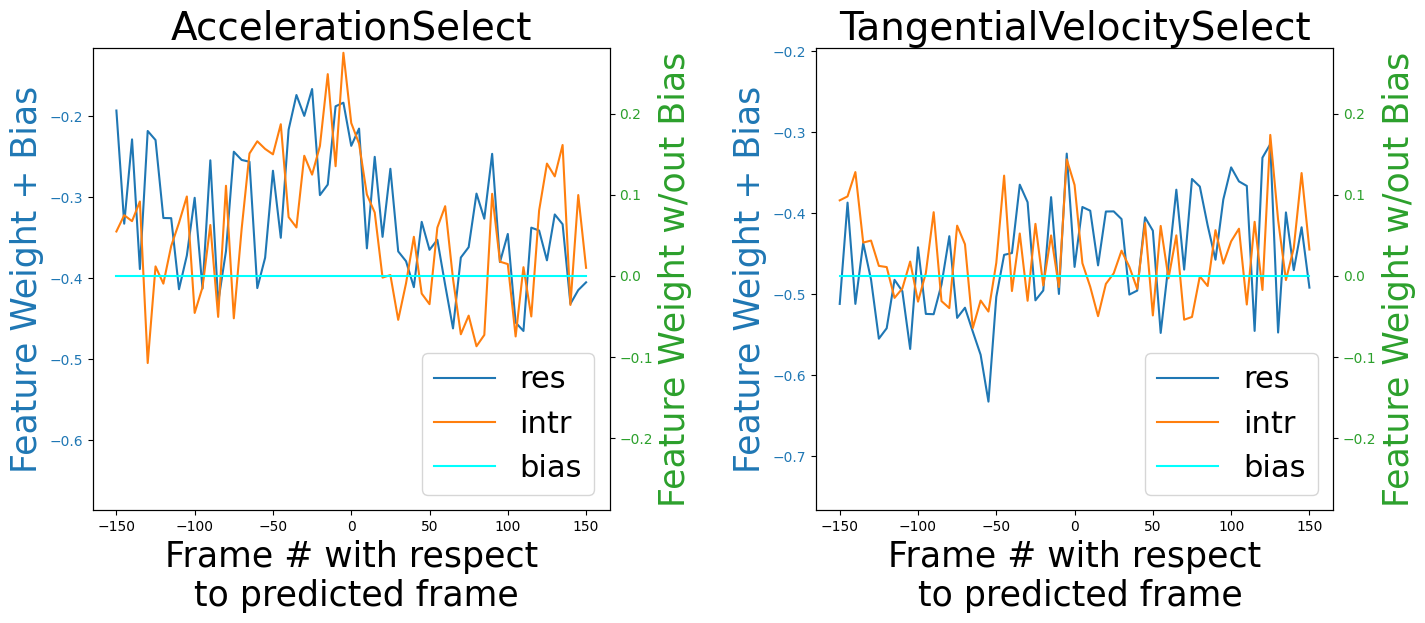}}\hspace{\fighspacer}  \\
& &  \\
\scriptsize{Program Filters}\hspace{\fighspacer} & \scriptsize{Decision Trees}\hspace{\fighspacer} &
\scriptsize{1D Conv Net}\hspace{\fighspacer} \\
 \includegraphics[width=\figsize\textwidth]{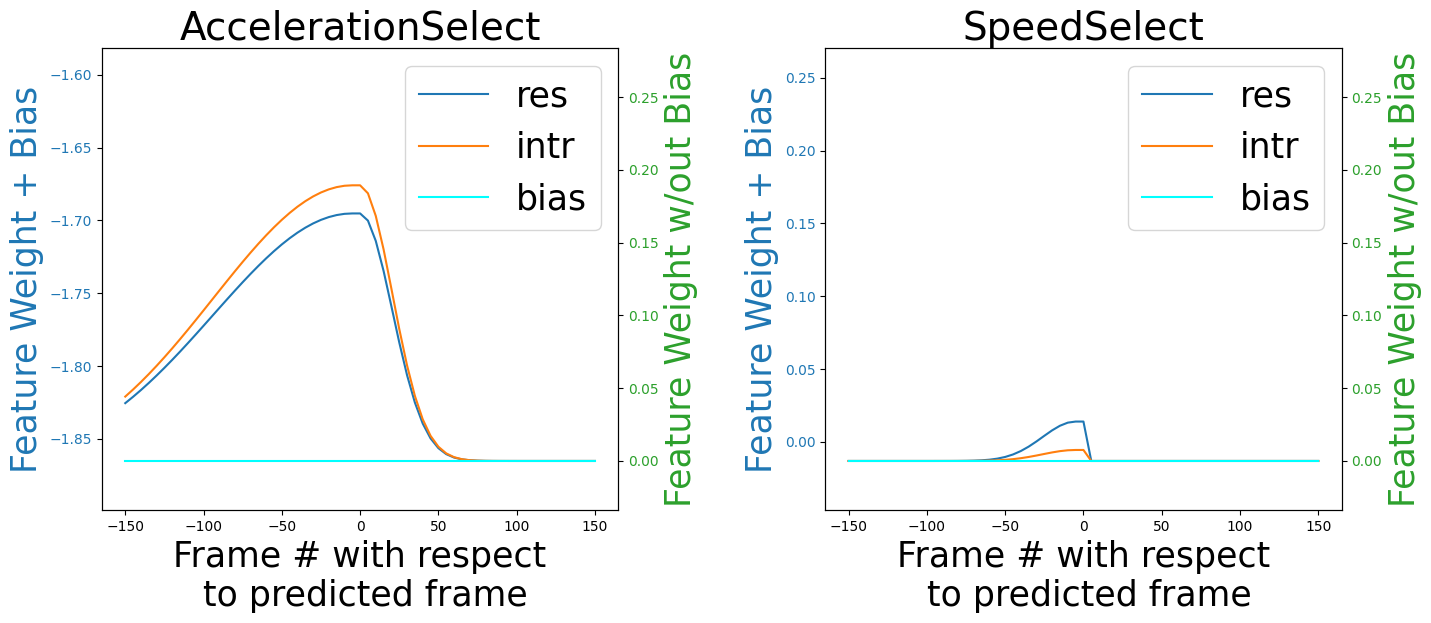}\hspace{\fighspacer} &
 \includegraphics[width=\figsize\textwidth]{{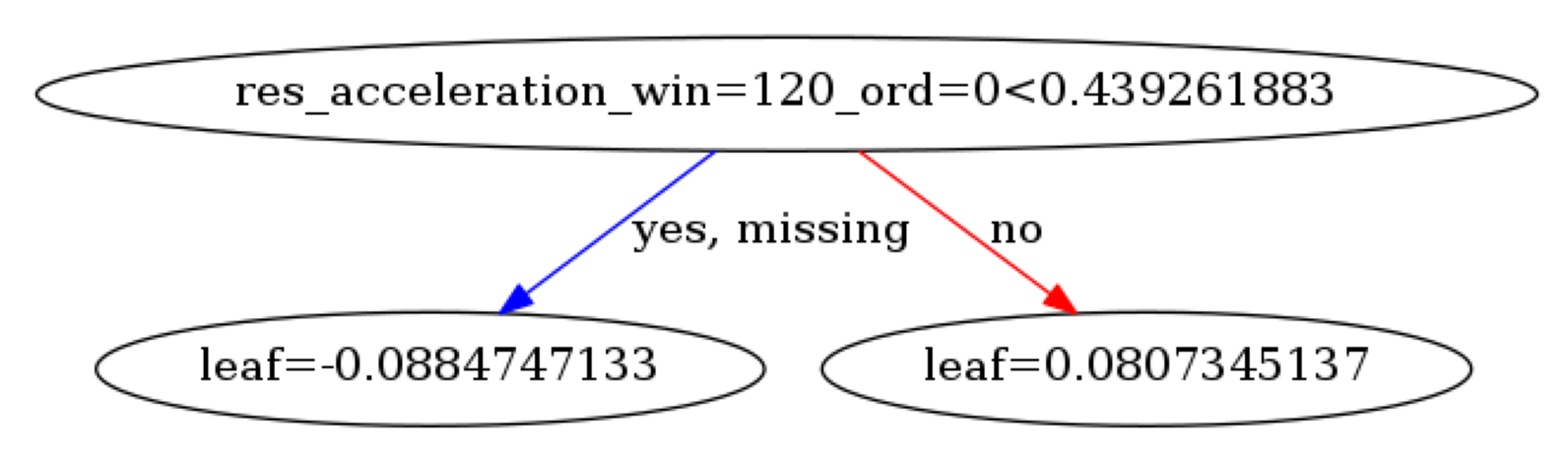}}\hspace{\fighspacer}  &
 \includegraphics[width=\figsize\textwidth]{{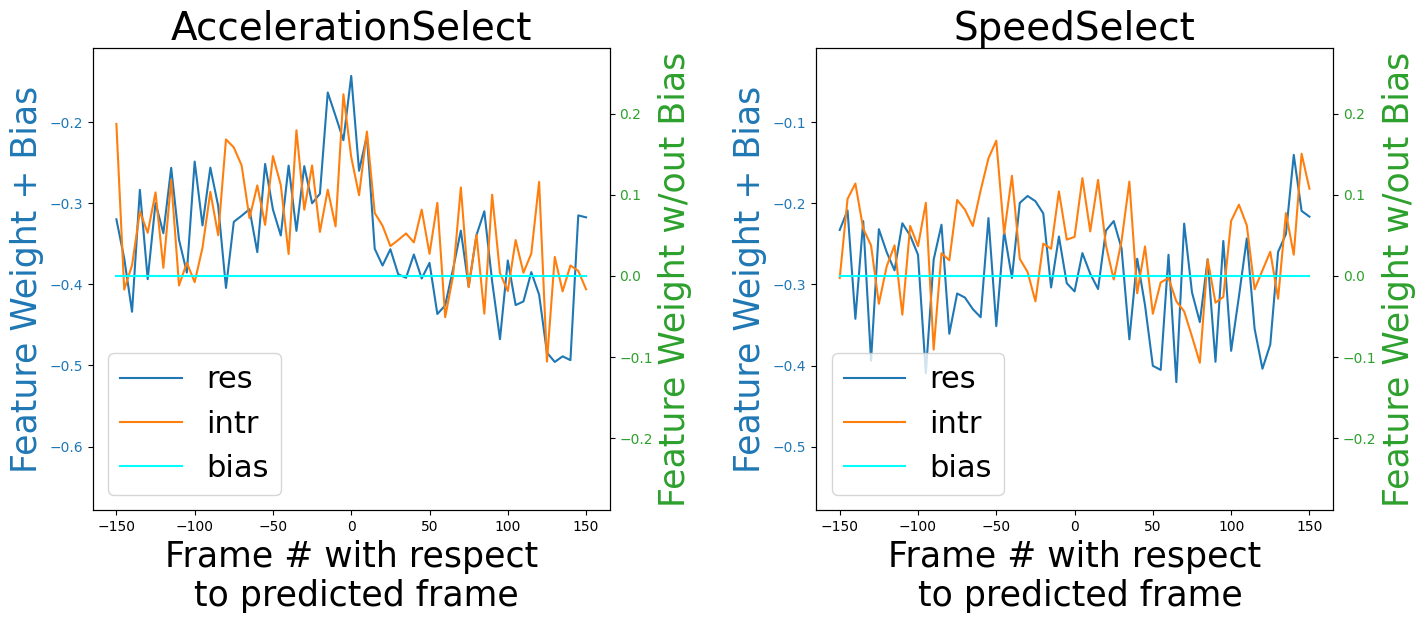}}\hspace{\fighspacer} \\

\end{tabular}
\caption{\textbf{Comparing models for two annotators.} Each row represents the visualized model trained on aggression vs. nonaggression annotations for one annotator. \emph{Left:} The program filters are from the learned disjunctions, and shows the weight applied at each timestamp for normalized trajectory features from program synthesis. \emph{Center:} Depth 1 decision tree with branches. \emph{Right:} Neural network weights on a subset of input features, matching each annotator's disjunction features.}
\label{fig:interpretability}
\end{figure*}

\textbf{Accuracy.} Synthesized programs with a disjunction of two filters achieve the highest F1 score for detection of interaction, and are comparable to the Decision Tree (DT) for detection of aggression (Figure~\ref{fig:100_bar}). Programs with a single filter had slightly lower F1 scores compared to the disjunction. For the DTs, the single depth 1 DT is much simpler than 10 depth 5 DTs. Single DT performs better on aggression, which implies that thresholding on one feature is able to classify aggression accurately and the deeper DT is more prone to overfitting. On the other hand, a more complex DT is needed to perform better on interaction.

\begin{figure}
    \centering
    \includegraphics[width=3.25in, height=1.75in]{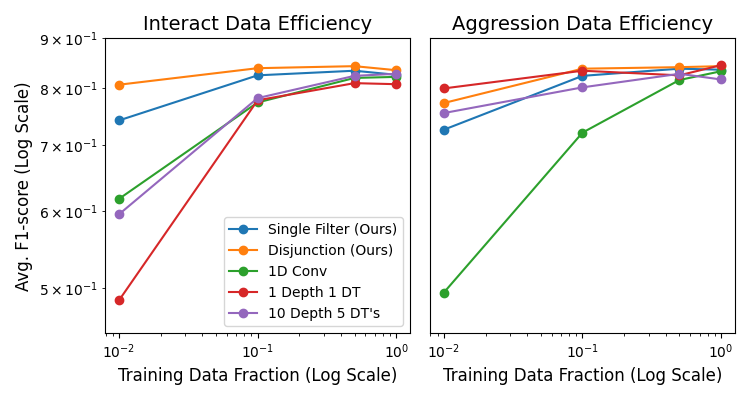}
    \caption{\textbf{Data Efficiency on Behavior Sequence Classification.} F1-score averaged across annotators vs. training data fraction on the aggression vs. non-aggression task (left) and interact vs. other task (right). }
    \label{fig:data_eff}
\end{figure}

In terms of data efficiency, disjunctions also remain the highest performing model on interact vs. other (Figure~\ref{fig:data_eff}). On aggression, disjunctions are comparable to the single depth 1 DT. Because of increased model complexity, the 1D Conv Net is generally less data efficient compared to our model. We verified that the variance in performance of both disjunctions and Morlet Filters are either less than or comparable to variance found in the baseline models.

\textbf{Interpretability.} We next visualized our models and baselines (Figure~\ref{fig:interpretability}). All three models include some aspect of temporal filtering of the data, however we argue that visualization and interpretation of this filtering is clearest for the disjunctions. The Conv Net filters appear as noisy versions of the disjunction filters, but without the disjunction filters as reference it is difficult for a domain expert to discern their structure. Filtering in the decision tree is implicit (in the names of the features used), and interpreting the numerical thresholds and leaf values is challenging. In contrast, the smoothness of the disjunction filters makes them easy to read, and their asymmetry around the predicted frame allows them to produce a variety of temporal structures. For domain experts to visualize our model more easily, we also added support for visualization of our trained models and their output within Bento~\cite{segalin2020mouse} (Figure~\ref{fig:bento_vis}).

\section{Conclusion}

\begin{figure}
    \centering
    \includegraphics[width=0.9\linewidth]{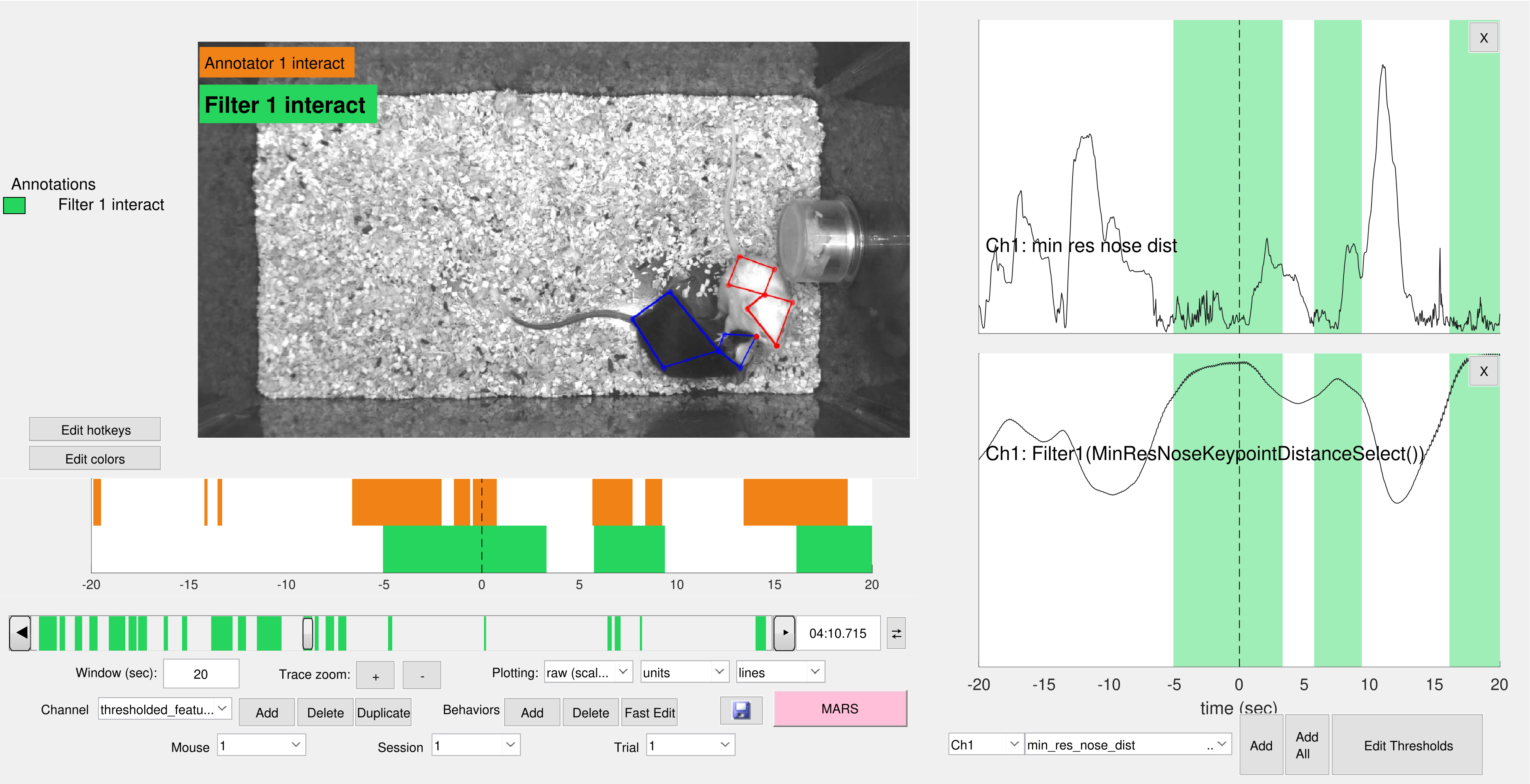}
    \caption{\textbf{Visualization of our learned filters in Bento~\cite{segalin2020mouse}}}
    \vspace{-0.1in}
    \label{fig:bento_vis}
\end{figure}

We propose a method, based on program synthesis, for learning programmatic descriptions of behavior annotations. We show that our method is accurate compared to baseline methods and that the programs we learn are qualitatively interpretable to domain experts. Automated behavior quantification systems for animal studies are often trained and evaluated on human-provided labels. As a result, human variability  will affect their performance. Programmatic explanations for annotated behavior can help us interpret annotation differences towards improving reproducibility of behavioral studies.

{\small
\bibliographystyle{ieee_fullname}
\bibliography{egbib}
}

\end{document}